\documentclass[english]{article}
\usepackage{babel}
\usepackage{amsmath}
\usepackage{amssymb}
\usepackage{graphicx}
\usepackage{authblk}

\usepackage[dvipsnames,table]{xcolor}
\usepackage[colorlinks]{hyperref}
\usepackage{multirow}
\usepackage{tabularx}
\usepackage[margin=1in]{geometry}

\renewenvironment{abstract}
{\begin{quote}
\noindent \rule{\linewidth}{.5pt}\par{}}
{\smallskip\noindent \rule{\linewidth}{.5pt}
\end{quote}
}

\begin{document}
\title{ULSA: Unified Language of Synthesis Actions for Representation of Synthesis Protocols}

\author[1,2,a]{Zheren Wang}
\author[1,2,a]{Kevin Cruse}
\author[1,2]{Yuxing Fei}
\author[1,c]{Ann Chia}
\author[2]{Yan Zeng}
\author[1,2]{Haoyan Huo}
\author[1,2]{Tanjin He}
\author[1,2]{Bowen Deng}
\author[1,*,b]{Olga Kononova}
\author[1,2,*]{Gerbrand Ceder}
\affil[1]{Department of Materials Science \& Engineering, University of California, Berkeley, CA 94720, USA}
\affil[2]{
Materials Sciences Division, Lawrence Berkeley National Laboratory, Berkeley, CA 94720, USA}
\affil[*]{Corresponding author: olga\_kononova@berkeley.edu and gceder@berkeley.edu}
\affil[a]{Equal contribution}
\affil[b]{Present address: Roivant Sciences, New York, NY 10036, USA}
\affil[c]{Present address: Nanyang Technological University, Republic of Singapore, 639798}

\renewcommand\Affilfont{\small}
\linespread{1.5}
\date{}
\maketitle


\newpage
\begin{abstract}
Applying AI power to predict syntheses of novel materials requires high-quality, large-scale datasets. 
Extraction of synthesis information from scientific publications is still challenging, especially for extracting synthesis actions, because of the lack of a comprehensive labeled dataset using a solid, robust, and well-established ontology for describing synthesis procedures.
In this work, we propose the first \textit{unified language of synthesis actions} (ULSA) for describing ceramics synthesis procedures. 
We created a dataset of 3,040 synthesis procedures annotated by domain experts according to the proposed ULSA scheme.
To demonstrate the capabilities of ULSA, we built
a neural network-based model to map arbitrary ceramics synthesis paragraphs into ULSA and used it to construct synthesis flowcharts for synthesis procedures. 
Analysis for the flowcharts showed that (a) ULSA covers essential vocabulary used by researchers when describing synthesis procedures and (b) it can capture important features of synthesis protocols. 
This work is an important step towards creating a synthesis ontology and a solid foundation for autonomous robotic synthesis. 

\end{abstract}

\newpage
\section{Introduction}

In the past decade, we have witnessed the growing success of data-driven and artificial intelligence (AI)-based methodologies promoting breakthroughs in predicting materials structure, properties, and functionality \cite{Alberi_2018, HimanenDDMSPerspective, SchmidtMLMSReview}. 
Nonetheless, adapting the power of AI to predict and control materials synthesis and fabrication is still challenging and requires substantial effort in gathering high-quality large-scale datasets. 
One approach to gather such datasets of synthesis parameters and conditions would be running high-throughput experiments. 
This requires a costly setup and substantial human labor and expertise, and is typically limited to a small part of chemical space. 
Another way to acquire the data or augment existing datasets is to extract information about materials synthesis from the wealth of scientific publications (e.g. papers, archives, patents) available online. 

Scientific text mining has received its recognition in the past few years \cite{KononovaISci,OlivettiAPR20, IRreview}, providing the materials science community with datasets on a variety of materials and their properties \cite{huang2020database, court2018auto, court2020magnetic} as well as synthesis protocols \cite{kim2017machine, kononova2019text, Mysore2019}.
Nonetheless, a majority of these text mining studies have been focused on extracting chemical entities such as material names, formulas, properties, and other characteristics \cite{Eltyeb2014, swain2016chemdataextractor, Jessop2011a, westonJCIM2019, hiszpanski2020nanomaterials}.
There have only been a few attempts to extract information about chemical synthesis and reactions and compile them into the flowchart of synthesis actions \cite{Hawizy2011, Mysore2019, kuniyoshi2020annotating, Vaucher2020}.
This is largely due to the lack of comprehensive labeled datasets or annotation schema needed to train algorithms. 
Indeed, publicly available large-scale collections of standardized labeled data for named entities recognition (NER) tasks are well established in the biochemical and biomedical domains (GENIA \cite{KimGENIA}, CHEMDNER \cite{KrallingerCHEMDNER2015}).
Materials science datasets are less standardized and mainly task-specific \cite{dieb2015framework, Kulkarni2018,friedrich2020sofcexp}.
To the best of our knowledge, the only publicly available annotated corpus of materials synthesis protocols was published by Mysore et al. \cite{Mysore2019}. 
It contains 230 labeled synthesis paragraphs with labels assigned to material entities, synthesis actions, and other synthesis attributes. 

A major obstacle in annotating synthesis actions in the text corpora is the lack of a solid, robust, and well-established ontology for describing synthesis procedures in materials science \cite{Ontology}.
Indeed, researchers prefer to vaguely sketch ``methods'' sections of the manuscript in general human-readable language rather than follow a specific protocol.
This significantly impacts reproducibility of the results, not to mention ambiguity in understanding even when read by a human expert \cite{Ontology}. 
While such ambiguity is inconvenient for human readers, the growing interest in automated AI-guided materials synthesis demands a robust and unified language for describing synthesis protocols in order to make them applicable to autonomous robotic platforms \cite{SzymanskiMatHor, HammerJACS, MehrScience2020}.

\begin{figure}[h]
\centering
\includegraphics[width=0.95\linewidth]{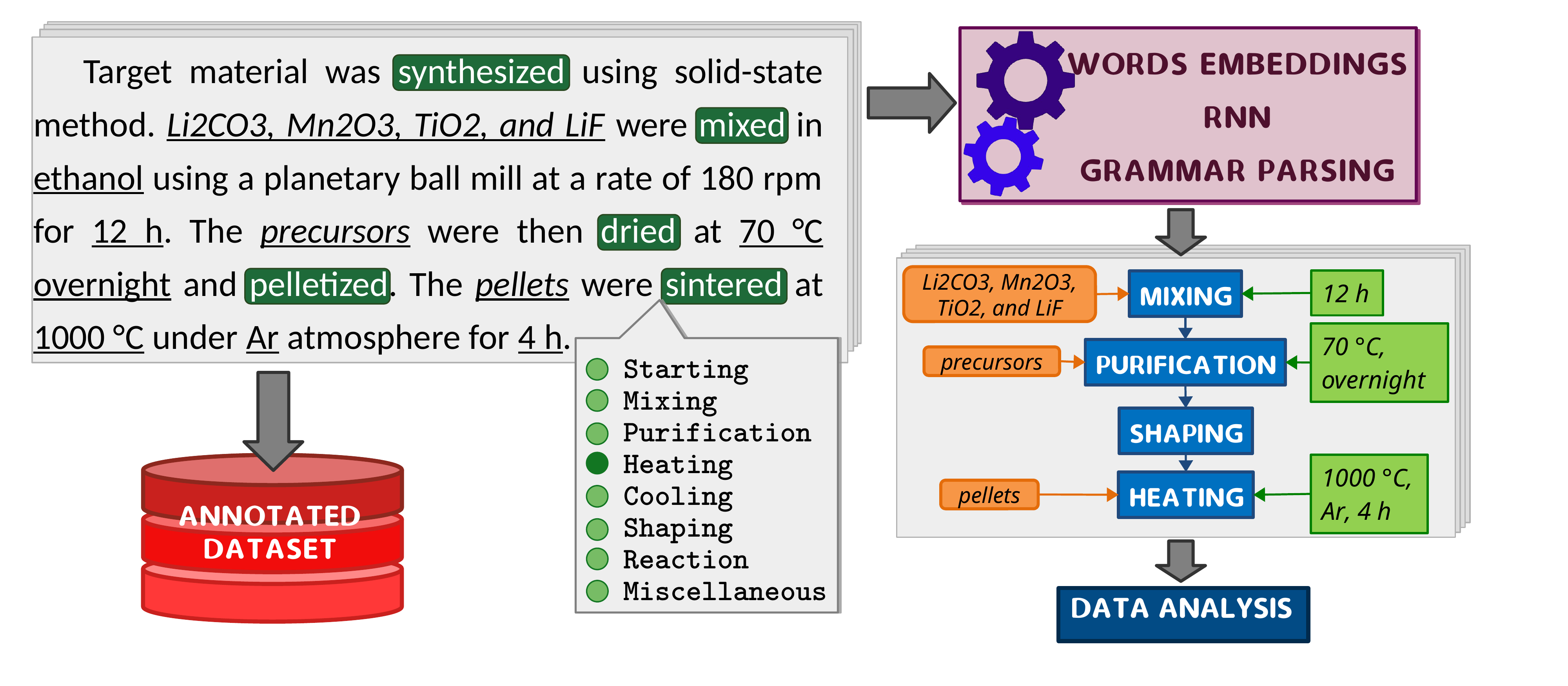}
\caption{
\textbf{Schematic workflow of data annotation, extraction and analysis.} 
First, the set of paragraphs were annotated using an Amazon Mechanical Turk engine.
Highlighted in green are the action token that were annotated and then extracted using a neural network model.
Other highlighted tokens and phrases (i.e. synthesis action attributes and subjects) were obtained using rule-based sentence parsing solely for the purpose of data analysis and are not presented in the annotated dataset.
The obtained labeled dataset is stored as single JSON file and is also used for training a neural network model to identify synthesis actions in the text.
Obtained synthesis actions, attributes and subjects were converted into synthesis flowcharts that was further used for data analysis. 
}
\label{fig:intro}
\end{figure}

In this work, we propose a \textit{unified language of synthesis actions} (ULSA) to describe solid-state, sol-gel, precipitation, and solvo-/hydrothermal synthesis procedures.
We also present a labeled dataset of 3,040 synthesis sentences created using the proposed ULSA schema. 
To verify applicability of the ULSA and the dataset, we trained a neural network-based model that identifies a sequence of synthesis actions in a paragraph, maps them into the ULSA, and builds a graph of the synthesis procedure (Figure~\ref{fig:intro}).
Analysis of the graphs from thousands of paragraphs has shown that this ULSA vocabulary is large enough to obtain high-accuracy extraction of synthesis actions as well as to pick the important features of each of the aforementioned synthesis types.
The dataset and the script for building such a synthesis flowchart is publicly available. 
We anticipate that  these results will be widely used by the researchers interested in scientific text mining and will help to achieve a breakthrough in predictive and AI-guided autonomous materials synthesis.

\section{Methodology}
\subsection{Unified Language of Synthesis Actions and annotation scheme}
\label{sec:Annotation}
To unify terminology used to describe a synthesis procedure, we defined 8 \textit{action terms} that unambiguously identify a type of synthesis action.
Every action word (or multi-word phrase) in the dataset is mapped to the corresponding action term according to the following rule: the word (or multi-word phrase) is recognized as an action if it (a) results in modification of the state of the material or mixture during the synthesis or (b) carries a piece of information affecting the outcome of the synthesis procedure.
The action terms used within the unified language are explained below.
In each example, the text underlined is the word or phrase that is annotated.

\begin{itemize}
    \item \texttt{Starting}: A word or a multi-word phrase that marks the beginning of a synthesis procedure. 
    Specifically, this often indicates which materials will be produced. 
    For example: 
    \textit{``PMN-PT was \underline{synthesized} by the columbite precursor method''}, 
    \textit{``Solid-state synthesis was \underline{used to prepare} the target material''},
    \textit{``The powder was \underline{obtained} after the aforementioned procedure''}.
    \item \texttt{Mixing}: A word or a multi-word phrase that marks the combination of different materials (in a solid or liquid phase) to form one substance or mass.
    For example: 
    \textit{``Precursors were weighted and \underline{ball -milled}''},
    \textit{``Precursors were \underline{mixed} in appropriate amounts''},
    \textit{``Sb\textsubscript{2}O\textsubscript{3} is \underline{added} to the solution''}, 
    \textit{``The solution was \underline{neutralized}''}, 
    \textit{``The mixture was \underline{stabilized} by the addition of sodium citrate''}.
    \item \texttt{Purification}: A word or a multi-word phrase that marks the separation of the sample phases. 
    This also includes drying of a material.
    For example: 
    \textit{``Samples were \underline{exfoliated} from substrates''},
    \textit{``The liquid was discarded and the remaining product was \underline{filtered off} and \underline{washed} several times with distilled water''}, 
    \textit{``The precursors were \underline{heated} in order to remove the moisture''}, 
    \textit{``The precipitation was \underline{collected} by washing the solution in distilled water''}.
    \item \texttt{Heating}: A word or a multi-word phrase that marks increasing or maintaining high temperature for the purpose of obtaining a specific sample phase or promoting a reaction rather than drying a sample.
    For example: 
    \textit{``The powder sample was \underline{annealed} to obtain a crystalline phase''},
    \textit{``The mixture was subjected to \underline{heating} at 240 °C for 24 h''}.
    \item \texttt{Cooling}: A word or a multi-word phrase that marks rapid, regular, or slow cooling of a sample.
    For example:
    \textit{``The product was \underline{cooled} down to room temperature in the furnace''},
    \textit{``The sample was \underline{quenched} rapidly in the solid CO\textsubscript{2}''},
    \textit{``The products was \underline{left to cool} down to room temperature''}.
    \item \texttt{Shaping}: A word or a multi-word phrase that marks the compression of powder or forming the sample to a specific shape.
    For example:
    \textit{``The powder was \underline{pressed} into circular pellets''},
    \textit{``The powder was then \underline{pelletized} with a uniaxial press''}.
    \item \texttt{Reaction}: A word or a multi-word phrase that marks a transformation without any external action. 
    For example:
    \textit{``The sample was \underline{left to react} for 6 hrs''},
    \textit{``The temperature was \underline{kept} at 1000 K''},
    \textit{``The solution was \underline{maintained} at 200 K for 12 hrs''}.
    \item \texttt{Miscellaneous:} A word or a multi-word phrase that marks an action done on a sample that either does not induce any transformation of the sample or does not belong to any of the above classes.
    \textit{``The pellets were \underline{placed} in a sealed alumina crucible''},
    \textit{``The reaction vessel was \underline{wrapped} with aluminum foil''},
    \textit{``The sample was \underline{sealed} in a tube''},
    \textit{``The gel was \underline{transferred} to an oven''}.
\end{itemize}

\subsection{Dataset annotation}
To annotate synthesis paragraphs with the unified language of synthesis actions (ULSA), we selected 535 synthesis paragraphs from the database of 420K full-text publications acquired previously \cite{kononova2019text}.
The paragraphs where chosen to proportionally represent four major types of ceramics synthesis: solid-state, sol-gel, solvo-/hydrothermal, and precipitation.
The details of the content acquisition and synthesis type classification have been described in previous papers \cite{kononova2019text, huo2019semi}.

The 535 paragraphs consisted of 3,781 tokenized sentences \cite{swain2016chemdataextractor}. 
First, each sentence was classified as either related to synthesis or not related to synthesis.
The latter case usually contains sentences about product characterization and other details. 
Next, we isolated 3,040 synthesis sentences and assigned labels to each word or multi-word phrase in the sentence on the basis of the ULSA protocol with annotation schema described in Section~\ref{sec:Annotation}. 
Only words and phrases describing synthesis actions were annotated. The final dataset consists of these 3,040 labeled synthesis sentences.
All annotations were performed using a custom Amazon Mechanical Turk-based server.

\subsection{Annotation decisions and ambiguous cases}

The ULSA was developed based on the authors' own experiences with the extraction of information from materials synthesis paragraphs \cite{kononova2019text} and extensive communication with experimentalists actively involved in various types of materials synthesis research. 
The annotation schema and the choice of action terms were designed to provide maximum flexibility to future users and allow them to adjust the schema according their preferences and tasks.
For example, the annotated multi-word phrases such as ``subjected to heating'', ``left to react'', and ``heated to evaporate'' were handled as one entity. 
This way, they can be split into individual terms or modified later with a simple set of rules to make a customized labeled dataset.

It is important to keep in mind that we mapped words into the terms of synthesis action per sentence, meaning that we used only information in the context of a given sentence to make a decision about the annotation of a word, rather than the whole paragraph. 
The reason for this choice is the multiple and diverse possibilities to combine and augment sentences leading to different meanings of the terms. 
The interpretation of the whole text or paragraph is an entirely separate field of research that is outside the scope of this work. 

We chose to annotate those words that are characteristic of a synthesis procedure or result in the transformation of a substance.
In other words, those actions which are usually performed by default are not annotated.
For example, in the sentence ``the solution was sealed in an autoclave'', no terms would be annotated as actions since the sealing step for hydrothermal synthesis is considered a default step.
Similarly, in the sentence ``the precursors were weighed and mixed,'' the term ``weighed'' is not a synthesis action since it is to be expected in synthesis, while ``mixing'' is a synthesis action because it may have a specific condition and transform the sample, or can be preceded by calcination of the precursors in other syntheses.

The exclusion from this rule is the \texttt{Starting} action.
Even terms belonging to this action do not bring any special information or explicit action to the synthesis, we chose to distinguish ``starting'' actions because in a substantial number of cases they can serve as flags to separate multiple synthesis procedures from one another.
An illustration of this situation is when precursors are prepared prior to synthesizing a target material, as in sol-gel synthesis. 

For the annotation of \texttt{Mixing} synthesis actions, we did not differentiate between powder mixing, ball milling (grinding), addition of droplets, or dissolving of substances.
In many situations, this precise definition depends on the solubility of reactants and mixing environment, as well as on other details of the procedure that are never explicitly mentioned in the text. 
We leave it up to a user to create their own application-based definitions of these mixing categories. 
Nonetheless, in the application below we provide a rule-based example of how these types of synthesis actions can be identified in the text.

The \texttt{Miscellaneous} action term was introduced to make room for those synthesis actions that are not typical or do not fall into any other category but nevertheless appear as a synthesis action within our definitions. 
While \texttt{Miscellaneous} action terms can be easily confused with \texttt{Reaction} actions or non-actions, the decision depends on the sentence context and can be arbitrarily extended or removed by a user.
Comparing ``the sample was kept in the cruicible'' and ``the sample was kept overnight,''
the former is not a synthesis action while the latter should be considered an important synthesis step. 

Ambiguous situations as in the ones mentioned above are ubiquitous in descriptions of syntheses. 
A substantial amount of these situations occur when authors try to be wordy or use flowery language when writing the synthesis methods.
Unfortunately, this often presents a challenge for accurate machine interpretation of the text.
We accounted for some of these cases when annotated the data as described below.

First, implicit mentions of synthesis actions (i.e. when a past participle form of a verb is used as a descriptive adjective referring to an already processed material) is the most frequent source of confusion. 
We chose to annotate these as synthesis actions. 
For example: ``the \underline{calcined} powder was \underline{pressed} and \underline{annealed}.''
In this sentence, the descriptive adjective \textit{calcined} could be either a restatement of the fact that there was a calcination step or it could be additional information which had not been mentioned previously.
These situations can be later resolved with a rule-based approach, hence we leave it as a task for users of the data.

The situation when a method is specified along with the synthesis action is also common. 
In a phrase of the form ``transformed by a specific procedure,'' we consider only the key action (the transformation) as a synthesis action.
For example: ``the precipitates were \underline{separated} by centrifugation.'' 
When required, the method can be retrieved with a set of simple rules.
    
Redundant action phrases are also abundant in many descriptions of the procedures.
In a phrase of form ``subjected to a process'', we considered only the processing verb as a synthesis action.
For example: ``the samples were subjected to an initial \underline{calcination} process.''
    
Finally, phrases that attempt to reason the purpose of the action, such as ``left to react'', ``brought to a boil'', ``heated to evaporate,'' are considered as one synthesis action. 
This is done for the purpose of providing flexibility to a user and to let them make a decision on how to treat these cases.

\subsection{Synthesis terms mapping}
\label{sec:MappingMeth}
We used lookup table (baseline) and neural network models to map synthesis sentences into the ULSA.

\subsubsection{Baseline model}
\label{sec:Baseline}
Two baseline models were implemented, both based on a lookup table.
For the lookup table, we chose the most frequent words used to describe synthesis steps in the ``methods'' section of the papers.
The first baseline model matches every token against the lookup table and assigns the corresponding action term if any appear.
The second baseline model uses information about the part of speech of a given word (assigned by SpaCy \cite{SpaCy}) and matches only verbs against the lookup table.

\subsubsection{Word embeddings}
Word embeddings were used as a vectorized representation of the word tokens for the neural network model.
To create an embedding, we trained a Word2Vec model \cite{mikolov2013distributed} implemented in the Gensim library \cite{gensim}. 
We used $\sim$420K paragraphs describing four synthesis types: solid-state, sol-gel, solvo-/hydrothermal and precipitation synthesis.
The paragraphs were obtained as described in our previous work \cite{kononova2019text}.
Prior to training, the text was normalized and tokenized using ChemDataExtractor \cite{swain2016chemdataextractor}.
Conjunctive adverbs describing consequences, such as ``therefore'', ``whereas'', and ``next'', were removed from the text.
All quantity tokens were replaced with a keyword \texttt{<NUM>}, and all chemical formulas were replaced with keyword \texttt{<CHEM>}.
All words that occur less than 5 times in the text corpus were replaced with the keyword \texttt{<UNK>}.
We found that skip-gram with negative sampling loss (n = 10) performed best, and the final embedding dimension was set to 100. 

\subsubsection{Neural network model}
\label{sec:LSTM}
We used a bi-directional long short-term memory (bi-LSTM) neural network model to map synthesis tokens into the aforementioned action terms. 
The model was implemented using the Keras library (\href{https://keras.io/}{https://keras.io/}) with latent dimensionality 32 and dropout probability 0.2.
Word embeddings were used as model input.
The categorical cross-entropy was calculated as the loss function. The labeled dataset was split into training, test, and validation sets using a 70:20:10 split, respectively.
Early stopping was used to obtain the best performance.

\subsection{Data analysis}
\subsubsection{Reassignment of mixing terms}
\label{sec:Reassign}
For data analysis, we separated \texttt{Mixing} synthesis action terms into \texttt{Dispersion Mixing} and \texttt{Solution Mixing} whenever there was enough information to distinguish between the two, otherwise they were left as \texttt{Mixing} action.
Here, \texttt{Dispersion Mixing} is identified either by explicit ``dispersion'' action words or by words such as ``grinding'' or ``milling'' plus any liquid environment.
\texttt{Solution Mixing} is identified by a list of specific action words such as ``dissolve'', ``dropwise added'', and others.
For this, we constructed and traversed the dependency trees of the sentences using SpaCy library \cite{SpaCy} and used dictionaries of common solution and mixing terms.

\subsubsection{Constructing synthesis flowchart for paragraphs}
\label{sec:GraphMeth}
For every paragraph in the set, we then applied the bi-LSTM mapping model (Section~\ref{sec:MappingMeth}) to extract the sequence of action terms from every sentence.
Next, we merged all the synthesis actions obtained from all sentences within the paragraph into a synthesis flowchart.
This was performed with a rule-based approach by traversing grammar trees and analysing the surrounding words of each action term and comparing them to the words and action terms of the previous sentence.
Finally, the flowchart of synthesis actions for a given paragraph was converted into an adjacency matrix.
For this, synthesis action terms were ordered and assigned to rows and columns of the matrix and initialized with zeros, resulting in a 10 by 10 matrix for every paragraph (8 action terms from vocabulary of ULSA plus two additional terms for \texttt{Mixing} term).
Whenever there was a step from action $i$ to action $j$, the corresponding value in the matrix was incremented by 1.
The matrices for all paragraphs were flattened and merged together for further principal component analysis.

\section{Results}

\subsection{Code and data availability}

The dataset of 3,040 annotated synthesis sentences as well as the processing scripts are available \newline at CederGroupHub/synthesis-action-retriever at \url{https://doi.org/10.5281/zenodo.5644302}.
In the dataset, each record contains the raw sentence tokens concatenated with a space between each token and a list of objects, each containing a token and the tag assigned to that token.
For example:
\begin{verbatim}
    {
        "annotations" : 
            [
                {
                    "tag" : token_tag,
                    "token" : token 
                }
            ],
        "sentence" : sentence
    }
\end{verbatim}

The repository also contains a script for training a bi-LSTM model that can be used to map words into action terms. 
Users are not limited to using only the provided dataset, but can augment their usage with other labeled data as long as they satisfy the data format described above. 
Finally, we also share scripts used for the inference of synthesis actions terms and for building synthesis flowcharts for a list of paragraphs. 
Examples of model application are available as well. 

\subsection{Dataset statistics}
\begin{figure}[t]
\centering
\includegraphics[width=0.95\linewidth]{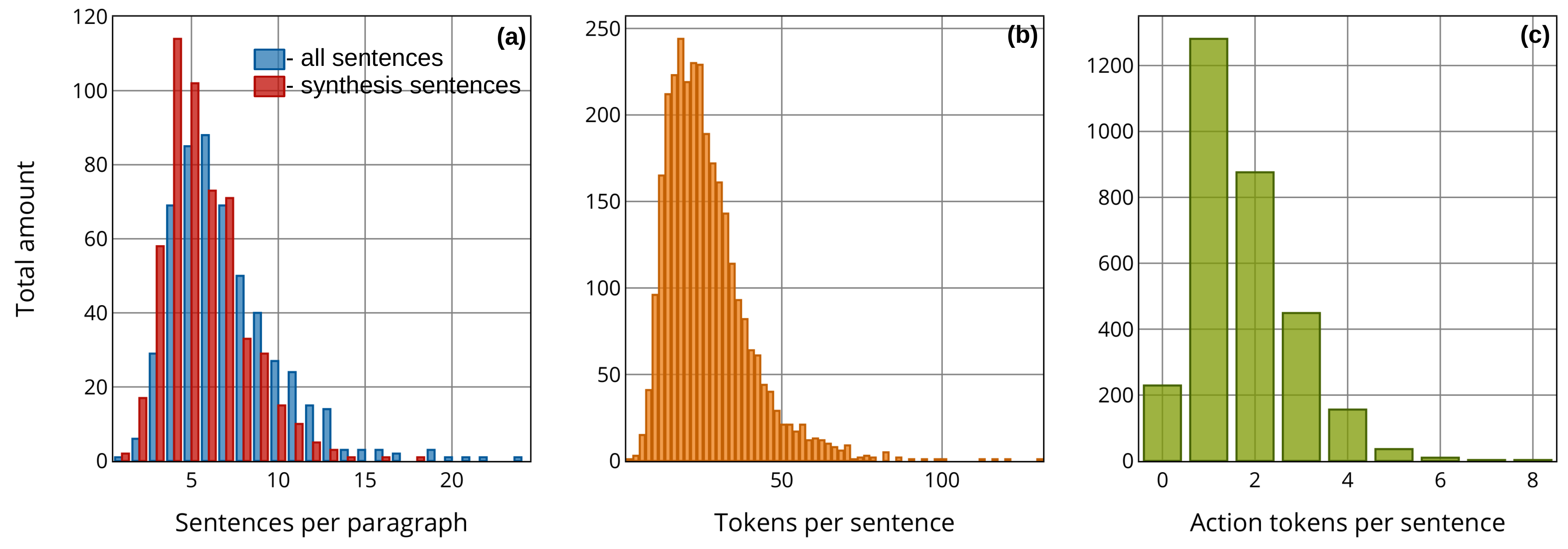}
\caption{
\textbf{Qualitative characteristics of the annotated dataset.}
(a): Number of sentences per paragraph (blue), including sentences related to synthesis procedure (red).
(b): Number of all tokens per sentence in the annotated set.
(c): Number of tokens denoting a synthesis action per sentence in the annotated set.  
}
\label{fig:hist}
\end{figure}

The quantitative characteristics of the set are provided in Table~\ref{tab:stat} and displayed in Figure~\ref{fig:hist}.
Briefly, 535 synthesis paragraphs resulted in 3,781 sentences of which 3,040 describe actual synthesis procedures.
While we tried to maintain an even distribution of the action terms in the labeled set, it is still highly skewed toward \texttt{Mixing} and \texttt{Purification} actions.
This is not surprising, since mixing of precursors occupies any synthesis procedure and purification is required in almost any non-solid-state method for ceramics synthesis. 
\texttt{Heating} is the next most prevalent synthesis action since it is also one of the basic operations in ceramic synthesis.

To probe the robustness of ULSA and our annotation schema, we asked 6 human experts to annotate the same paragraphs in our dataset and used Fleiss' kappa score to estimate the inter-annotator agreement between the annotations \cite{FleissKappa}.
In general, the Fleiss' kappa score evaluates the degree from -1 to 1 to which different annotators agree with one another above the agreement expected by pure chance. 
A positive Fleiss' kappa indicates good agreement, scores close to zero indicates near randomness in categorization, and negative scores indicate conflicting annotations. 
This is a generalized reliability metric and is useful for agreement between three or more annotators across three or more categories.

Table \ref{tab:fleiss} lists the Fleiss' kappa scores for agreement between human experts annotating the sentences according to the schema described in Section~\ref{sec:Annotation}.  
The table shows good agreement on distinguishing synthesis sentences from non-synthesis sentences, as well as for all and for each individual synthesis action, including non-actions. 
The agreement across all action terms is 0.83. 
Among those, the action terms with lower scores are \texttt{Shaping} and \texttt{Miscellaneous}. 
The low score for \texttt{Miscellaneous} is expected since a wide range of actions which do not induce a transformation in the sample could be mapped into this category. 
The \texttt{Shaping} action term can also be associated with many synthesis operations.
For instance, granulating procedures that break a sample into smaller chunks could be considered a \texttt{Shaping} action; at the same time, a bench chemist could consider ``granulation'' to be \texttt{Mixing} action term since it requires performing a grinding operation to obtain the new shape.
Less ambiguous actions terms, such as \texttt{Heating} and \texttt{Mixing}, showed higher agreement. 

\subsection{Mapping synthesis procedures into a unified language of synthesis actions}

\subsubsection{Mapping model}
As a first approach for mapping of synthesis paragraphs into ULSA, we used dictionary lookup constructed as described in Section~\ref{sec:Baseline}.
We use the labeled dataset of 3,040 sentences to assess the performance of the model.
We considered two options: mapping of all sentence words and mapping the verbs only. 
In both cases, the overall accuracy of the prediction (i.e. F1 score) is $\sim$60-70\% (Table~\ref{tab:accuracy}).
Nonetheless, mapping of all words shows relatively good recall and poor precision, while mapping of only verbs improves the precision but diminishes recall. 

These results moved us toward considering a recurrent neural network model for mapping paragraphs into ULSA.
The bi-LSTM model combined with word embeddings  (Section~\ref{sec:LSTM}) was trained on the labeled dataset of 3,040 sentences.
The bi-LSTM model significantly improves mapping accuracy, yielding $>$90\% F1 score. 
It is important to notice here that all the metrics for baseline and neural network models were computed per sentence, i.e. we evaluated the whole sentence being mapped correctly rather than individual terms. 

There are a few reasons why the bi-LSTM model outperforms plain dictionary lookup. 
First, researchers use diverse vocabulary to describe synthesis procedures, hence there are unlimited possibilities in constructing a lookup table. 
For instance, ``heating'' can be referred as ``calcining'', ``sintering'', ``firing'', ``burning'', ``heat treatment'', and so on.
In this case, a word embedding model helps to significantly improve the score even for those terms that have never appeared in training set (e.g. ``degas'', ``triturate'').
Second, a given verb is defined as a synthesis action term largely based on the context. 
Prominent examples are ``heating rate'', ``mixing environment'', ``ground powder'', etc.
That is well captured by the recurrent neural network architecture. 
Lastly, synthesis actions are not only denoted by verb tokens, but also by nouns, adjectives, and gerunds. 
This can be also learnt by the neural network better than by a set of rules. 

In summary, we designed a neural network-based model that maps any synthesis paragraph into ULSA with high accuracy and significantly outperforms a plain dictionary lookup approach. 

\subsubsection{Analysis of action embeddings}
\begin{figure}[t]
\centering
\includegraphics[width=0.9\linewidth]{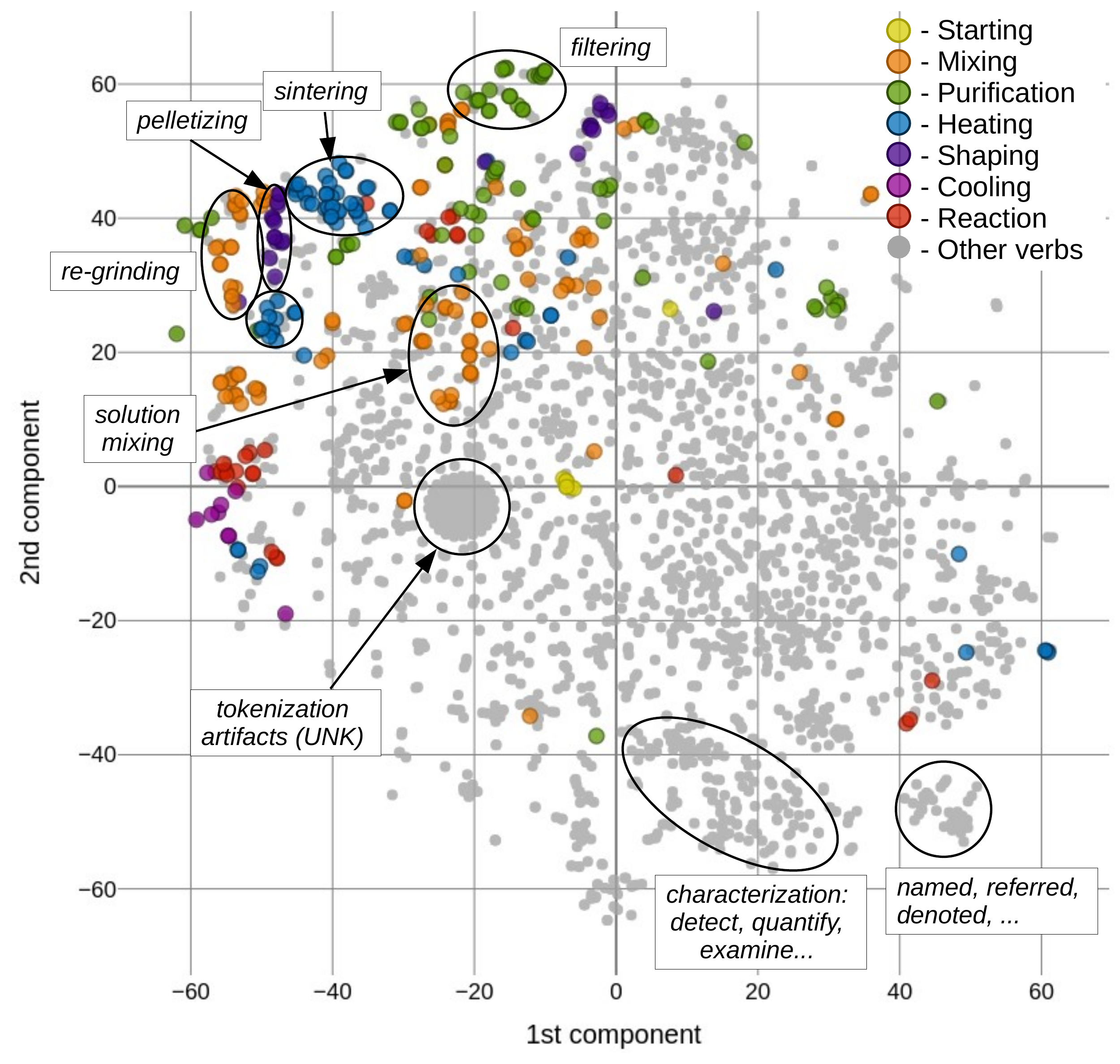}
\caption{
\textbf{2D projection of word embeddings vectors.} 
Shown are the most frequent verb tokens encountered in the set of 420K paragraphs describing a synthesis procedure.
Highlighted in different colors are the vectors that correspond to the common verbs from the categories of synthesis actions used for annotation.  
Other prominent clusters of vectors are denoted with circles and labeled by a common term.
Dimensionality reduction was performed using t-SNE approach.
}
\label{fig:embedd}
\end{figure}

To analyse how well the ULSA represents the space of synthesis operations commonly used when describing ceramics synthesis processes, we plotted a 2D projection of the word embeddings calculated with a t-SNE approach.
The results are shown in Figure~\ref{fig:embedd}.
To achieve a clear representation, we only analysed those verbs that appear more than 10 times.
We then mapped these paragraphs into ULSA by using the bi-LSTM model.
Those verbs that were assigned with a ULSA label are color-coded in the figure correspondingly, the other non-synthesis action terms are colored in grey.

First, we observe that the verbs mapped into ULSA and hence representing synthesis actions are all grouped in the top-left corner of the projection.
Indeed, analysis of the individual words in the rest of the space showed that those are the words that generally appear in synthesis paragraphs but do not carry any information about the synthesis procedure. 
For instance, these are verbs denoting characterization of a material (``detect'', ``quantify'', ``examine'', ``measure''), naming of a sample (``denoted'', ``referred'', ``named'', ``labeled'') or referring to a table or figure.
The blob of dots in the middle of the plot are all words that were either mis-tokenized during text segmentation or mistakenly recognized as verbs by the SpaCy algorithm. 
In the embeddings mapping, these words are replaced with the \texttt{<UNK>} token. 

A second interesting observation is that the embeddings of firing (blue dots), pelletizing  (purple dots) and grinding into powder (orange dots) are all located next to each other.
This agrees well with the fact that those actions together describe solid-state synthesis processes.
Oppositely, the verbs describing solution mixing (orange dots) are in close proximity with the verbs referring to purification or drying (green dots).
Similarly, verbs indicating cooling processes (magenta dots) and the verbs referring to reaction processes (red dots) are clustered together.
This agrees with the often encountered constructions of ``left to cool'' or ``kept and then cooled'' describing the final steps of a given synthesis.

Taken together, these results demonstrate that 
(a) the embeddings model we created reflects well the similarity of the verbs used for synthesis descriptions and
(b) the vocabulary of ULSA covers all common synthesis actions used in ceramics synthesis. 

\subsubsection{Analysis of graphs clustering}
As we showed above, ULSA can capture well the vocabulary commonly used for the description of synthesis and, further, we were able to design a high-accuracy model that maps arbitrary synthesis descriptions into ULSA. 
However, we want also to make sure that unification of synthesis actions still allows for distinguishing between ceramics synthesis types. 
For that purpose, we constructed synthesis flowcharts for 4,000 paragraphs (1,000 per each synthesis type) randomly pulled from the set of 420K ceramics synthesis paragraph (see Section~\ref{sec:GraphMeth} for procedure description). 
For constructing the flowchart for a synthesis (represented by an adjacency matrix), we used the synthesis action terms assigned to each sentence in a paragraph.
Additionally, we augmented \texttt{Mixing} actions with two categories, \texttt{Dispersion Mixing} and \texttt{Solution Mixing}, by using heuristics and dictionary lookup (Section~\ref{sec:Reassign}). 
It is important to note here that we assume a linear order of synthesis actions, i.e. that the sequence of sentences and synthesis actions in a paragraph corresponds to the true sequence of synthesis steps done during experiment. 
According to our estimation, this assumption is violated only in 2\% of paragraphs in the 420K paragraphs set. 

All the adjacency matrices were flattened and concatenated, resulting in a matrix of size 100$\times$4000, i.e. 10$\times$10 matrix per each of 4,000 paragraphs, where 10 is the size of the ULSA vocabulary with two additional mixing actions.
Next, principal component analysis was used to perform dimensionality reduction of the matrix. 

\begin{figure}[t]
\centering
\includegraphics[width=0.9\linewidth]{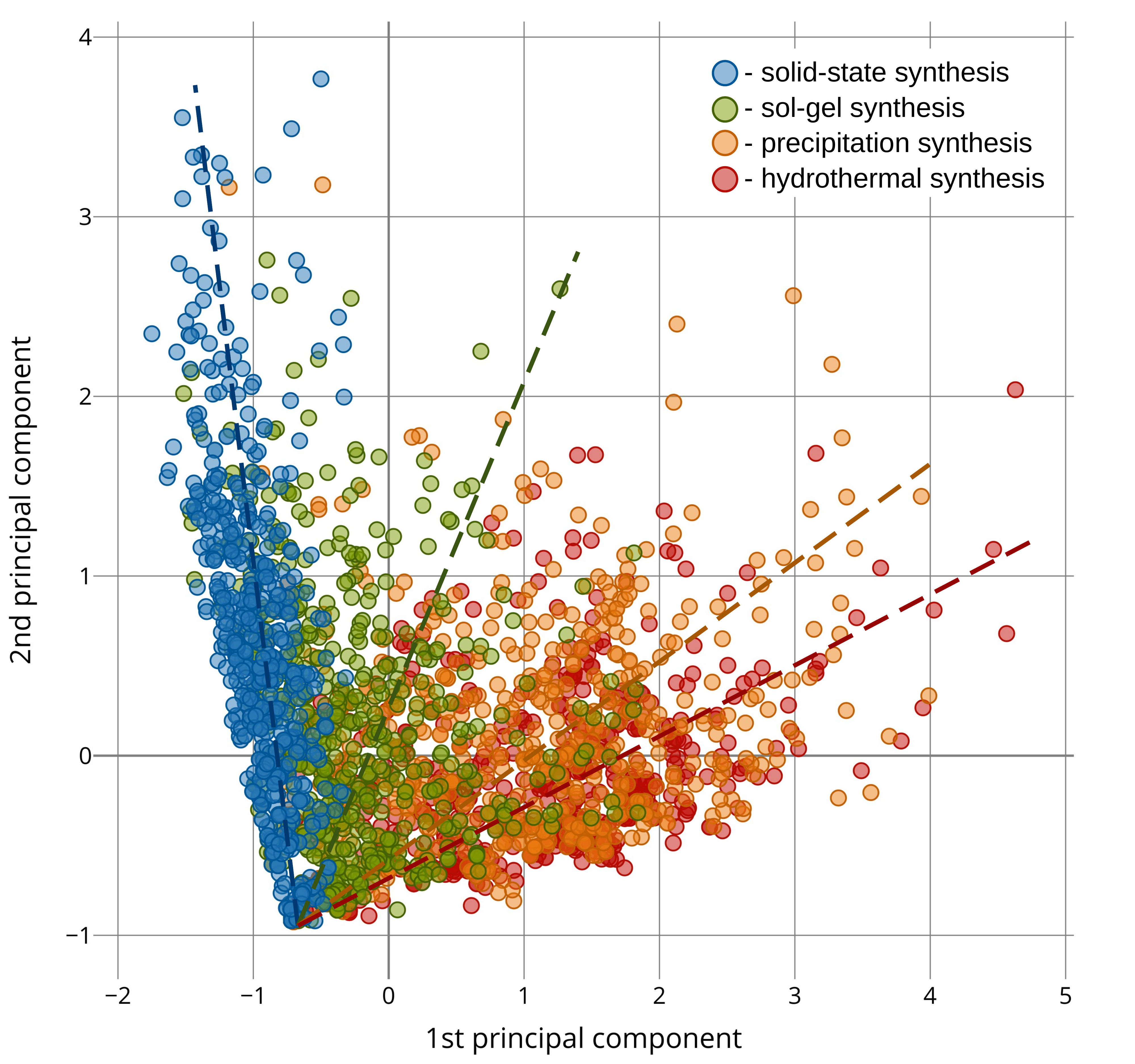}
\caption{
\textbf{Visualization of the first two principal components for the adjacency matrices of synthesis action graphs.} 
Each dot on the plot represent a synthesis graph colored according to its type. 
Dash lines display linear fitting of each data subset and show the overall direction for clustering of each synthesis graph. 
Note that the lines were shifted to have a common origin for representation purposes while preserving the slope. 
}
\label{fig:clustering}
\end{figure}

Figure~\ref{fig:clustering} displays the projection of the 1st and 2nd principal components for each synthesis flowchart with different colors corresponding to different types of syntheses. 
A few observations can be made from the plot.
First, the data points corresponding to solid-state synthesis are narrowly clustered along a line with negative slope unlike the other synthesis types which are spread widely and whose linear fittings have positive inclination.
Second, the clusters of data points for precipitation and hydrothermal synthesis almost completely overlap and partially overlap with sol-gel synthesis, while overlapping with solid-state synthesis is negligible. 

These two observations agree well with the standard procedures associated with each of the four synthesis types.
Indeed, solid-state syntheses usually operate with mixing powder precursors, firing the mixture, and obtaining final products; sol-gel synthesis is considered as a solid-state synthesis with solution-assisted mixing of precursors; hydrothermal and precipitation syntheses usually involve preparation of the sample in solution, then filtering (purification) to separate the liquid and obtain the final product instead of including a firing step.

To get further insights, we sampled and compared synthesis procedures along each of the fitted lines.
The results show that the 1st principle component correlates with the involvement of solution mixing for precursors.
In other words, the larger and more positive the data point along the 1st principle component, the more steps of dissolving and mixing precursors in solution as well as purification that data point involves.
This agrees well with the fact that solid-state synthesis mostly operates with powders while hydrothermal and precipitation procedures are solution-based procedures, and sol-gel syntheses exist in between.

The 2nd principal component corresponds to the level of complexity of the syntheses procedure. 
The larger and more positive the data point along the 2nd principle component, the more synthesis steps become involved in the process. 
Interestingly, all four synthesis types exhibit simple synthesis procedures (fewer steps) and complex synthesis procedures (many steps).
Nonetheless, solid-state synthesis has the largest deviation compared to hydrothermal and precipitation synthesis since solid-state procedures can involve multiple heating and re-grinding steps for the sample to obtain the desired material phase while in solution synthesis this can often be achieved in one or two steps.
 
\section{Discussion and Conclusions}
In this work, we aim to fill the gap in automated synthesis information extraction from scientific publications by proposing a unified language for synthesis actions (ULSA).
We used the ULSA on an annotated set of 3,040 sentences about ceramics synthesis including solid-state, sol-gel, precipitation and solvo-/hydrothermal syntheses.
The dataset is publicly available and can be easily customized by researchers accordingly to fit their application.
As an example of such application, we used a recurrent neural network and grammar parsing to build a mapping model that converts written synthesis procedures into a ULSA-based synthesis flowchart.
Analysis of the results demonstrates that the ULSA vocabulary spans the essential set of words used by researchers to describe synthesis procedures in scientific literature and that the flowchart representation of synthesis constructed using ULSA can capture important synthesis features and distinguish between solid-state, sol-gel, precipitation and solvo-/hydrothermal synthesis methods.

Despite these promising results, the ULSA scheme still suffers from imperfections and can be significantly improved in the future. 
First, we only demonstrated that it works for ceramics synthesis, and synthesis techniques such as deposition, crystal growth, and others may require extending the ULSA vocabulary or reconsidering the definitions of some terms. 
Second, the scheme and methodology will benefit from a robust approach to distinguish between various mixing procedures.
This includes separation between, for example, dissolving precursors and dispersive mixing in a liquid environment, using ball-milling to homogenize the sample and using high-energy ball-milling to actually achieve the final product, adding reagents to promote reaction and adding precursors to compensate for loss due to volatility, and other cases.
We have demonstrated that the details of mixing are important for distinguishing between ceramics synthesis methods using simple heuristics, however, the scheme will benefit from a high-fidelity approach. 
Nonetheless, we anticipate that our results and the ULSA schema will help researchers to develop a data-oriented methodology to predict synthesis routes of novel materials.

Efficient and controllable materials synthesis is a bottleneck in technological breakthroughs. 
While predicting materials with advanced properties and functionality has been brought to a state-of-the-art level with the development of computational and data-driven approaches, the design and optimization of synthesis routes for those materials is still a tedious experimental task. 
The progress in inorganic materials synthesis is mainly impeded due to (a) lack of publicly available large-scale repositories with high-quality synthesis data and (b) lack of ontology and standardization for communication on synthesis protocols. 
Indeed, the first matter arises from the fact that the vast majority of experimental data gets buried in lab notebooks and is never published anywhere.
As a result, researchers are liable to perform redundant and wasteful experimental screenings through those parameters of synthesis that have already been performed by someone, but are not reported.
Even published experimental procedures face the problem of ambiguity of the language used by researchers. 
This creates a major challenge in acquiring synthesis data from publications by automated approaches including text mining. 

The advantage of the paradigm we establish in this work is that it brings us closer to addressing important questions in materials synthesis: \textit{``How should we think about the synthesis process?''}, \textit{``What is the minimum information required to unambiguously identify a synthesis procedure?''}, and \textit{``Can synthesis be thought of as a combination of fixed action blocks augmented with attributes such as temperature, time, and environment, or are there other important aspects that have to be taken into account?''}. 
These questions will become crucial when transitioning towards AI-driven synthesis.

Recent developments in autonomous robotic synthesis and the attempts to ``close the feedback loop'' in making decisions for the next synthesis step make the question of synthesis ontology and unification especially important \cite{SzymanskiMatHor, Burger2019Nature, HammerJACS}.
Indeed, while theoretical decision-making and AI-guided systems can operate with abstract synthesis representations, implementation of this methodology to an autonomous robotic platform will require well-defined and robust mapping onto a fixed set of manipulations and devices available to the robot.
The unified language we propose in this work can become a solid foundation for the future development in this direction. 

\section*{Author Contributions}
Z.W., K.C. O.K. and G.C. conceived the idea, and drafted the manuscript. 
Z.W., K.C., A.C. and O.K. implemented the algorithms and analyzed the data.
Z.W., Y.F., and H.H built the annotation tool.
Z.W., K.C., Y.F., Y.Z., and O.K. defined the annotation schema.
Z.W., K.C, Y.F., H.H., T.H., and B.D. prepared the annotation dataset.
All authors discussed and revised the manuscript.

\section*{Conflicts of interest}
There are no conflicts to declare.

\section*{Acknowledgements}
The authors would like to thank the team of librarians from the University of California, Berkeley: Anna Sackmann (Data Services Librarian), Rachael Samberg (Scholarly Communication Officer) and Timothy Vollmer (Scholarly Communication \& Copyright Librarian) for helping us to navigate through publishers copyright policies and related issues. 
We also thank Prof. Wenhao Sun (University of Michigan) for helpful discussions and thoughts about materials synthesis. 

This work was primarily supported 
by the National Science Foundation under Grant No. DMR-1922372, 
by the U.S. Department of Energy, Office of Science, Basic Energy Sciences, Materials Sciences and Engineering Division under Contract No. DE-AC02-05-CH11231 within the GENESIS EFRC program (DE-SC0019212).
Expert validation of the extracted data was supported by the Assistant Secretary of Energy Efficiency and Renewable Energy, Vehicle Technologies Office, U.S. Department of Energy under Contract No. DE-AC02-05CH11231.

\newpage
\section*{Tables}

\begin{table}[ht]
\centering
\begin{tabular}{|l|r|}
\hline
\rowcolor{YellowGreen}
 & \textbf{Amount} \\
\hline
Paragraphs used for annotation & 535 \\
\hline
Per synthesis type: & \\
 -- solid-state synthesis & 199 \\
 -- sol-gel synthesis & 51 \\
 -- solvo-/ hydrothermal synthesis & 148 \\
 -- precipitation & 137 \\
\hline
Total sentences & 3781\\
\hline
Synthesis sentences & 3040\\
\hline
Action tokens & 5547\\
\hline
Per action category: & \\
 -- starting & 619\\
 -- mixing & 1853\\
 -- purification & 1080\\
 -- heating & 973\\
 -- cooling & 259\\
 -- shaping & 225\\
 -- reaction & 232\\
 -- miscellaneous & 306\\
\hline
\end{tabular}
\caption{
Quantitative characteristics of the dataset chosen for annotation with ULSA schema.
}
\label{tab:stat}
\end{table}

\begin{table}[ht]
\centering
\begin{tabular}{|l|r|}
\hline
\rowcolor{YellowGreen}
 & \textbf{Score} \\
\hline
Identification of synthesis sentences & 0.69 \\
Action terms tagging &  0.83\\
\hline
Per action terms: & \\
 -- starting & 0.82\\
 -- mixing & 0.86\\
 -- purification & 0.79\\
 -- heating & 0.84\\
 -- cooling & 0.88\\
 -- shaping & 0.59\\
 -- reaction & 0.66\\
 -- miscellaneous & 0.45\\
 -- no action & 0.87 \\
\hline
\end{tabular}
\caption{
Fleiss' kappa score for inter-annotator agreement using ULSA scheme.
}
\label{tab:fleiss}
\end{table}

\begin{table}[ht]
\centering
\begin{tabular}{|l|c|c|c|}
\hline
\rowcolor{YellowGreen}
\textbf{Model} & \textbf{Precision} & \textbf{Recall}  & \textbf{F1 score}\\

Baseline 1 & 0.54 & 0.61 & 0.57 \\
 -- solid-state & 0.53  & 0.72 & 0.61 \\
 -- sol-gel & 0.57 & 0.75 & 0.65 \\
 -- hydrothermal & 0.54 & 0.53 & 0.54  \\
 -- precipitation & 0.55 & 0.50 & 0.53 \\
\hline

Baseline 2 & 0.84 & 0.50 & 0.63 \\
 -- solid-state & 0.84  & 0.54 & 0.66 \\
 -- sol-gel & 0.79 & 0.62 & 0.69 \\
 -- hydrothermal & 0.84 & 0.47 & 0.61 \\
 -- precipitation & 0.84 & 0.44 & 0.54 \\
\hline
bi-LSTM & 0.90  & 0.88 & 0.89 \\
 -- solid-state & 0.90  & 0.90 & 0.90 \\
 -- sol-gel & 0.88 & 0.86 & 0.87 \\
 -- hydrothermal & 0.90& 0.86 & 0.88  \\
 -- precipitation & 0.90  & 0.91 & 0.91 \\
\hline
\end{tabular}
\caption{
Performance of baseline and bi-LSTM models for mapping synthesis sentence into ULSA terms. 
In Baseline 1, all words in the sentence are matched against a lookup table.
In Baseline 2, only verbs tagged by SpaCy are matched against the lookup table.
The quantities are computed per sentence, i.e. the number of sentences with all the action tokens identified and assigned correctly.
}
\label{tab:accuracy}
\end{table}
\newpage
\section*{Figures}
\linespread{1.5}

\bibliographystyle{naturemag}
\bibliography{refs}

\begin{thebibliography}{10}
\expandafter\ifx\csname url\endcsname\relax
  \def\url#1{\texttt{#1}}\fi
\expandafter\ifx\csname urlprefix\endcsname\relax\def\urlprefix{URL }\fi
\providecommand{\bibinfo}[2]{#2}
\providecommand{\eprint}[2][]{\url{#2}}

\bibitem{Alberi_2018}
\bibinfo{author}{Alberi, K.} \emph{et~al.}
\newblock \bibinfo{title}{The 2019 materials by design roadmap}.
\newblock \emph{\bibinfo{journal}{J. Phys. D: Appl. Phys}}
  \textbf{\bibinfo{volume}{52}}, \bibinfo{pages}{013001}
  (\bibinfo{year}{2018}).

\bibitem{HimanenDDMSPerspective}
\bibinfo{author}{Himanen, L.}, \bibinfo{author}{Geurts, A.},
  \bibinfo{author}{Foster, A.} \& \bibinfo{author}{Rinke, P.}
\newblock \bibinfo{title}{Data-driven materials science: Status, challenges,
  and perspectives}.
\newblock \emph{\bibinfo{journal}{Advanced Science}}
  \textbf{\bibinfo{volume}{6}} (\bibinfo{year}{2019}).

\bibitem{SchmidtMLMSReview}
\bibinfo{author}{Schmidt, J.}, \bibinfo{author}{Marques, M.},
  \bibinfo{author}{Botti, S.} \& \bibinfo{author}{Marques, M.}
\newblock \bibinfo{title}{Recent advances and applications of machine learning
  in solid-state materials science}.
\newblock \emph{\bibinfo{journal}{npj Computational Materials}}
  \textbf{\bibinfo{volume}{5}} (\bibinfo{year}{2019}).

\bibitem{KononovaISci}
\bibinfo{author}{Kononova, O.} \emph{et~al.}
\newblock \bibinfo{title}{Opportunities and challenges of text mining in
  materials research}.
\newblock \emph{\bibinfo{journal}{iScience}} \textbf{\bibinfo{volume}{24}},
  \bibinfo{pages}{102155} (\bibinfo{year}{2021}).

\bibitem{OlivettiAPR20}
\bibinfo{author}{Olivetti, E.} \emph{et~al.}
\newblock \bibinfo{title}{Data-driven materials research enabled by natural
  language processing}.
\newblock \emph{\bibinfo{journal}{Appl. Phys. Rev.}}
  \textbf{\bibinfo{volume}{7}}, \bibinfo{pages}{041317} (\bibinfo{year}{2020}).

\bibitem{IRreview}
\bibinfo{author}{Krallinger, M.}, \bibinfo{author}{Rabal, O.},
  \bibinfo{author}{Lourenço, A.}, \bibinfo{author}{Oyarzabal, J.} \&
  \bibinfo{author}{Valencia, A.}
\newblock \bibinfo{title}{Information retrieval and text mining technologies
  for chemistry}.
\newblock \emph{\bibinfo{journal}{Chem. Rev.}} \textbf{\bibinfo{volume}{117}},
  \bibinfo{pages}{7673--7761} (\bibinfo{year}{2017}).

\bibitem{huang2020database}
\bibinfo{author}{Huang, S.} \& \bibinfo{author}{Cole, J.~M.}
\newblock \bibinfo{title}{A database of battery materials auto-generated using
  chemdataextractor}.
\newblock \emph{\bibinfo{journal}{Sci. Data}} \textbf{\bibinfo{volume}{7}},
  \bibinfo{pages}{1--13} (\bibinfo{year}{2020}).

\bibitem{court2018auto}
\bibinfo{author}{Court, C.} \& \bibinfo{author}{Cole, J.~M.}
\newblock \bibinfo{title}{Auto-generated materials database of curie and
  n{\'e}el temperatures via semi-supervised relationship extraction}.
\newblock \emph{\bibinfo{journal}{Sci. Data}} \textbf{\bibinfo{volume}{5}},
  \bibinfo{pages}{180111} (\bibinfo{year}{2018}).

\bibitem{court2020magnetic}
\bibinfo{author}{Court, C.} \& \bibinfo{author}{Cole, J.}
\newblock \bibinfo{title}{Magnetic and superconducting phase diagrams and
  transition temperatures predicted using text mining and machine learning}.
\newblock \emph{\bibinfo{journal}{npj Comput. Mater}}
  \textbf{\bibinfo{volume}{6}}, \bibinfo{pages}{1--9} (\bibinfo{year}{2020}).

\bibitem{kim2017machine}
\bibinfo{author}{Kim, E.} \emph{et~al.}
\newblock \bibinfo{title}{Machine-learned and codified synthesis parameters of
  oxide materials}.
\newblock \emph{\bibinfo{journal}{Sci. Data}} \textbf{\bibinfo{volume}{4}},
  \bibinfo{pages}{170127} (\bibinfo{year}{2017}).

\bibitem{kononova2019text}
\bibinfo{author}{Kononova, O.} \emph{et~al.}
\newblock \bibinfo{title}{Text-mined dataset of inorganic materials synthesis
  recipes}.
\newblock \emph{\bibinfo{journal}{Sci. Data}} \textbf{\bibinfo{volume}{6}},
  \bibinfo{pages}{1--11} (\bibinfo{year}{2019}).

\bibitem{Mysore2019}
\bibinfo{author}{Mysore, S.} \emph{et~al.}
\newblock \bibinfo{title}{The materials science procedural text corpus:
  Annotating materials synthesis procedures with shallow semantic structures}.
\newblock \emph{\bibinfo{journal}{LAW 2019 - 13th Linguistic Annotation
  Workshop, Proceedings of the Workshop}} \bibinfo{pages}{56--64}
  (\bibinfo{year}{2019}).
\newblock \eprint{1905.06939}.

\bibitem{Eltyeb2014}
\bibinfo{author}{Eltyeb, S.} \& \bibinfo{author}{Salim, N.}
\newblock \bibinfo{title}{Chemical named entities recognition: A review on
  approaches and applications}.
\newblock \emph{\bibinfo{journal}{J. Cheminform.}}
  \textbf{\bibinfo{volume}{6}}, \bibinfo{pages}{1--12} (\bibinfo{year}{2014}).

\bibitem{swain2016chemdataextractor}
\bibinfo{author}{Swain, M.~C.} \& \bibinfo{author}{Cole, J.~M.}
\newblock \bibinfo{title}{Chemdataextractor: a toolkit for automated extraction
  of chemical information from the scientific literature}.
\newblock \emph{\bibinfo{journal}{J. Chem. Inf. Model.}}
  \textbf{\bibinfo{volume}{56}}, \bibinfo{pages}{1894--1904}
  (\bibinfo{year}{2016}).

\bibitem{Jessop2011a}
\bibinfo{author}{Jessop, D.~M.}, \bibinfo{author}{Adams, S.~E.},
  \bibinfo{author}{Willighagen, E.~L.}, \bibinfo{author}{Hawizy, L.} \&
  \bibinfo{author}{Murray-Rust, P.}
\newblock \bibinfo{title}{Oscar4: a flexible architecture for chemical
  text-mining}.
\newblock \emph{\bibinfo{journal}{J. Cheminform.}}
  \textbf{\bibinfo{volume}{3}}, \bibinfo{pages}{41} (\bibinfo{year}{2011}).

\bibitem{westonJCIM2019}
\bibinfo{author}{Weston, L.} \emph{et~al.}
\newblock \bibinfo{title}{Named entity recognition and normalization applied to
  large-scale information extraction from the materials science literature}.
\newblock \emph{\bibinfo{journal}{J. Chem. Inf. Model.}}
  \textbf{\bibinfo{volume}{59}}, \bibinfo{pages}{3692--3702}
  (\bibinfo{year}{2019}).

\bibitem{hiszpanski2020nanomaterials}
\bibinfo{author}{Hiszpanski, A.} \emph{et~al.}
\newblock \bibinfo{title}{Nanomaterials synthesis insights from machine
  learning of scientific articles by extracting, structuring, and visualizing
  knowledge}.
\newblock \emph{\bibinfo{journal}{J. Chem. Inf. Model.}}
  \textbf{\bibinfo{volume}{60}}, \bibinfo{pages}{2876–2887}
  (\bibinfo{year}{2020}).

\bibitem{Hawizy2011}
\bibinfo{author}{Hawizy, L.}, \bibinfo{author}{Jessop, D.~M.},
  \bibinfo{author}{Adams, N.} \& \bibinfo{author}{Murray-Rust, P.}
\newblock \bibinfo{title}{Chemicaltagger: A tool for semantic text-mining in
  chemistry}.
\newblock \emph{\bibinfo{journal}{J. Cheminform.}}
  \textbf{\bibinfo{volume}{3}}, \bibinfo{pages}{1--13} (\bibinfo{year}{2011}).

\bibitem{kuniyoshi2020annotating}
\bibinfo{author}{Kuniyoshi, F.}, \bibinfo{author}{Makino, K.},
  \bibinfo{author}{Ozawa, J.} \& \bibinfo{author}{Miwa, M.}
\newblock \bibinfo{title}{Annotating and extracting synthesis process of
  all-solid-state batteries from scientific literature} (\bibinfo{year}{2020}).
\newblock \eprint{2002.07339}.

\bibitem{Vaucher2020}
\bibinfo{author}{Vaucher, A.} \emph{et~al.}
\newblock \bibinfo{title}{Automated extraction of chemical synthesis actions
  from experimental procedures}.
\newblock \emph{\bibinfo{journal}{Nat. Commun.}} \textbf{\bibinfo{volume}{11}},
  \bibinfo{pages}{3601} (\bibinfo{year}{2020}).

\bibitem{KimGENIA}
\bibinfo{author}{Kim, J.-D.}, \bibinfo{author}{Ohta, T.},
  \bibinfo{author}{Tateisi, Y.} \& \bibinfo{author}{Tsujii, J.}
\newblock \bibinfo{title}{Genia corpus -- a semantically annotated corpus for
  bio-textmining}.
\newblock \emph{\bibinfo{journal}{Bioinformatics}}
  \textbf{\bibinfo{volume}{19}}, \bibinfo{pages}{i180--i182}
  (\bibinfo{year}{2003}).

\bibitem{KrallingerCHEMDNER2015}
\bibinfo{author}{Krallinger, M.} \emph{et~al.}
\newblock \bibinfo{title}{The chemdner corpus of chemicals and drugs and its
  annotation principles}.
\newblock \emph{\bibinfo{journal}{J. Cheminform.}}
  \textbf{\bibinfo{volume}{7}}, \bibinfo{pages}{S2} (\bibinfo{year}{2015}).

\bibitem{dieb2015framework}
\bibinfo{author}{Dieb, T.}, \bibinfo{author}{Yoshioka, M.},
  \bibinfo{author}{Hara, S.} \& \bibinfo{author}{Newton, M.}
\newblock \bibinfo{title}{Framework for automatic information extraction from
  research papers on nanocrystal devices}.
\newblock \emph{\bibinfo{journal}{Beilstein J. Nanotechnol.}}
  \textbf{\bibinfo{volume}{6}}, \bibinfo{pages}{1872--1882}
  (\bibinfo{year}{2015}).

\bibitem{Kulkarni2018}
\bibinfo{author}{Kulkarni, C.}, \bibinfo{author}{Xu, W.},
  \bibinfo{author}{Ritter, A.} \& \bibinfo{author}{Machiraju, R.}
\newblock \bibinfo{title}{An annotated corpus for machine reading of
  instructions in wet lab protocols}.
\newblock In \emph{\bibinfo{booktitle}{Proceedings of the 2018 Conference of
  the North American Chapter of the Association for Computational Linguistics:
  Human Language Technologies, Volume 2 (Short Papers)}},
  \bibinfo{pages}{97--106} (\bibinfo{publisher}{Association for Computational
  Linguistics}, \bibinfo{address}{Stroudsburg, PA, USA}, \bibinfo{year}{2018}).

\bibitem{friedrich2020sofcexp}
\bibinfo{author}{Friedrich, A.} \emph{et~al.}
\newblock \bibinfo{title}{The {SOFC}-exp corpus and neural approaches to
  information extraction in the materials science domain}.
\newblock In \emph{\bibinfo{booktitle}{Proceedings of the 58th Annual Meeting
  of the Association for Computational Linguistics}},
  \bibinfo{pages}{1255--1268} (\bibinfo{publisher}{Association for
  Computational Linguistics}, \bibinfo{year}{2020}).

\bibitem{Ontology}
\bibinfo{author}{Kim, E.}, \bibinfo{author}{Huang, K.},
  \bibinfo{author}{Kononova, O.}, \bibinfo{author}{Ceder, G.} \&
  \bibinfo{author}{Olivetti, E.}
\newblock \bibinfo{title}{Distilling a materials synthesis ontology}.
\newblock \emph{\bibinfo{journal}{Matter}} \textbf{\bibinfo{volume}{1}},
  \bibinfo{pages}{8--12} (\bibinfo{year}{2019}).

\bibitem{SzymanskiMatHor}
\bibinfo{author}{Szymanski, N.~J.} \emph{et~al.}
\newblock \bibinfo{title}{Toward autonomous design and synthesis of novel
  inorganic materials}.
\newblock \emph{\bibinfo{journal}{Mater. Horiz.}} \bibinfo{pages}{--}
  (\bibinfo{year}{2021}).

\bibitem{HammerJACS}
\bibinfo{author}{Hammer, A. J.~S.}, \bibinfo{author}{Leonov, A.~I.},
  \bibinfo{author}{Bell, N.~L.} \& \bibinfo{author}{Cronin, L.}
\newblock \bibinfo{title}{Chemputation and the standardization of chemical
  informatics}.
\newblock \emph{\bibinfo{journal}{JACS Au}} \textbf{\bibinfo{volume}{0}},
  \bibinfo{pages}{null} (\bibinfo{year}{0}).

\bibitem{MehrScience2020}
\bibinfo{author}{Mehr, S. H.~M.}, \bibinfo{author}{Craven, M.},
  \bibinfo{author}{Leonov, A.~I.}, \bibinfo{author}{Keenan, G.} \&
  \bibinfo{author}{Cronin, L.}
\newblock \bibinfo{title}{A universal system for digitization and automatic
  execution of the chemical synthesis literature}.
\newblock \emph{\bibinfo{journal}{Science}} \textbf{\bibinfo{volume}{370}},
  \bibinfo{pages}{101--108} (\bibinfo{year}{2020}).

\bibitem{huo2019semi}
\bibinfo{author}{Huo, H.} \emph{et~al.}
\newblock \bibinfo{title}{Semi-supervised machine-learning classification of
  materials synthesis procedures}.
\newblock \emph{\bibinfo{journal}{npj Comput. Mater}}
  \textbf{\bibinfo{volume}{5}}, \bibinfo{pages}{1--7} (\bibinfo{year}{2019}).

\bibitem{SpaCy}
\bibinfo{author}{Honnibal, M.} \& \bibinfo{author}{Johnson, M.}
\newblock \bibinfo{title}{An improved non-monotonic transition system for
  dependency parsing}.
\newblock In \emph{\bibinfo{booktitle}{Proceedings of the 2015 Conference on
  Empirical Methods in Natural Language Processing}},
  \bibinfo{pages}{1373--1378} (\bibinfo{publisher}{Association for
  Computational Linguistics}, \bibinfo{address}{Lisbon, Portugal},
  \bibinfo{year}{2015}).

\bibitem{mikolov2013distributed}
\bibinfo{author}{Mikolov, T.}, \bibinfo{author}{Sutskever, I.},
  \bibinfo{author}{Chen, K.}, \bibinfo{author}{Corrado, G.} \&
  \bibinfo{author}{Dean, J.}
\newblock \bibinfo{title}{Distributed representations of words and phrases and
  their compositionality} (\bibinfo{year}{2013}).
\newblock \eprint{1310.4546}.

\bibitem{gensim}
\bibinfo{author}{{\v R}eh{\r u}{\v r}ek, R.} \& \bibinfo{author}{Sojka, P.}
\newblock \bibinfo{title}{Software framework for topic modelling with large
  corpora}.
\newblock In \emph{\bibinfo{booktitle}{{Proceedings of the LREC 2010 Workshop
  on New Challenges for NLP Frameworks}}}, \bibinfo{pages}{45--50}
  (\bibinfo{publisher}{ELRA}, \bibinfo{address}{Valletta, Malta},
  \bibinfo{year}{2010}).

\bibitem{FleissKappa}
\bibinfo{author}{Fleiss, J.}
\newblock \bibinfo{title}{Measuring nominal scale agreement among many raters}.
\newblock \emph{\bibinfo{journal}{Psychological Bulletin}}
  \textbf{\bibinfo{volume}{76}}, \bibinfo{pages}{378--382}
  (\bibinfo{year}{1971}).

\bibitem{Burger2019Nature}
\bibinfo{author}{Burger, B.} \emph{et~al.}
\newblock \bibinfo{title}{A mobile robotic chemist}.
\newblock \emph{\bibinfo{journal}{Nature}} \textbf{\bibinfo{volume}{583}},
  \bibinfo{pages}{237--241} (\bibinfo{year}{2020}).

\end{thebibliography}

\end{document}